\documentclass[journal,twoside,web]{ieeecolor}
\usepackage{generic}
\usepackage{cite}
\usepackage{amsmath,amssymb,amsfonts}
\usepackage{algorithmic}
\usepackage{graphicx}
\usepackage{textcomp}
\usepackage{caption}
\usepackage{subcaption}
\usepackage{booktabs}
\usepackage{multirow}
\usepackage{xcolor}
\usepackage{pifont}
\usepackage[outdir=./]{epstopdf}
\newcommand{\cmark}{\ding{51}}%
\newcommand{\xmark}{\ding{55}}%
 
\def\BibTeX{{\rm B\kern-.05em{\sc i\kern-.025em b}\kern-.08em
    T\kern-.1667em\lower.7ex\hbox{E}\kern-.125emX}}
\begin{document}
\title{TinyAD: Memory-efficient anomaly detection for time series data in Industrial IoT}
\author{Yuting~Sun, Tong~Chen, \textit{Member, IEEE}, Quoc~Viet~Hung~Nguyen, and Hongzhi~Yin, \textit{Senior Member, IEEE}
\vspace{-0.5cm}
\thanks{Submitted for review on 2nd September, 2022. This work is partly sponsored by Australian Research Council under the streams of Future Fellowship (FT210100624),
Discovery Early Career Researcher Award (DE200101465,
DE230101033), Discovery Project (DP190101985), and UQ
New Staff Research Start-up Grant (NS-2103).}
\thanks{Y. Sun, T. Chen and H. Yin are with the School of Information Technology and Electrical Engineering, The University of Queensland, Australia (e-mail: yuting.sun@uqconnect.edu.au, tong.chen@uq.edu.au, h.yin1@uq.edu.au).}
\thanks{Q. V. H. Nguyen is with the Institute for Integrated and Intelligent Systems, Griffith University, Australia (e-mail: henry.nguyen@griffith.edu.au).}
\thanks{H. Yin is the corresponding author.}
}

\maketitle

\begin{abstract}
Monitoring and detecting abnormal events in cyber-physical systems is crucial to industrial production. With the prevalent deployment of the Industrial Internet of Things (IIoT), an enormous amount of time series data is collected to facilitate machine learning models for anomaly detection, and it is of the utmost importance to directly deploy the trained models on the IIoT devices. However, it is most challenging to deploy complex deep learning models such as Convolutional Neural Networks (CNNs) on these memory-constrained IIoT devices embedded with microcontrollers (MCUs).  To alleviate the memory constraints of MCUs, we propose a novel  framework named Tiny Anomaly Detection (TinyAD) to efficiently facilitate onboard inference of CNNs for real-time anomaly detection. First, we conduct a comprehensive analysis of depthwise separable CNNs and regular CNNs for anomaly detection and find that the depthwise separable convolution operation can reduce the model size by 50-90\% compared with the traditional CNNs. Then, to reduce the peak memory consumption of CNNs, we explore two complementary strategies, in-place, and patch-by-patch memory rescheduling, and integrate them into a unified framework.  The in-place method decreases the peak memory of the depthwise convolution by sparing a temporary buffer to transfer the activation results, while the patch-by-patch method further reduces the peak memory of layer-wise execution by slicing the input data into corresponding receptive fields and executing in order. Furthermore, by adjusting the dimension of convolution filters, these strategies apply to both univariate time series and multidomain time series features. Extensive experiments on real-world industrial datasets show that our framework can reduce peak memory consumption by 2-5x with negligible computation overhead.
\end{abstract}


\section{Introduction}
\label{sec:introduction}
The Industrial Internet of Things (IIoT) is a paradigm integrated with sensor devices and communication technologies to enhance productivity and automation in industrial systems (e.g., smart manufacturing and agriculture). However, the IIoT infrastructures supported by cyber-physical systems, such as smart water management systems and power grids, are susceptible to cyberattacks and sensor failure. Such vulnerability of IIoT systems suggests the importance of timely detection of abnormal events. By deploying anomaly detection techniques, these systems can be alerted to early changes in normal status and potential loss of data reliability. 

In the conventional IIoT network, the sensor data are normally transmitted through the IIoT network and fed into an expert cloud platform to leverage data-driven models for anomaly detection. However, the remote central computing unit can result in high latency due to communication overhead and resource scheduling\cite{s21248320}. Hence, a low-latency solution is to offload prediction models from the cloud servers to IIoT devices. Although the advancement in chip technology facilitates the capability of pushing machine intelligence to IIoT devices, deploying deep learning models to microcontroller (MCU) based devices is still in its nascent stage. Unlike mobile devices and edge devices with rich compute and storage resources (even GPUs),  most MCU-based IIoT devices (such as various industrial sensors) have internal Flash memory $\leq$1MB for data storage and SRAM $\leq$64kB for in-memory processing\cite{w14030309, 0_21}. This memory constraint of MCU hinders the on-device deployment of the deep learning-based anomaly detection methods, such as DeepAnT\cite{2018.2886457} and LSTM-AD\cite{Pankaj_Malhotra}. Thus, we aim to develop a solution for the state-of-the-art (SOTA) deep learning-based anomaly detection models to reach minimal memory consumption (especially peak memory) while preserving the prediction performance. 

\begin{figure*}[hbt!]
  \centering
  \captionsetup{justification=centering}
  \includegraphics[width= 0.8\linewidth]{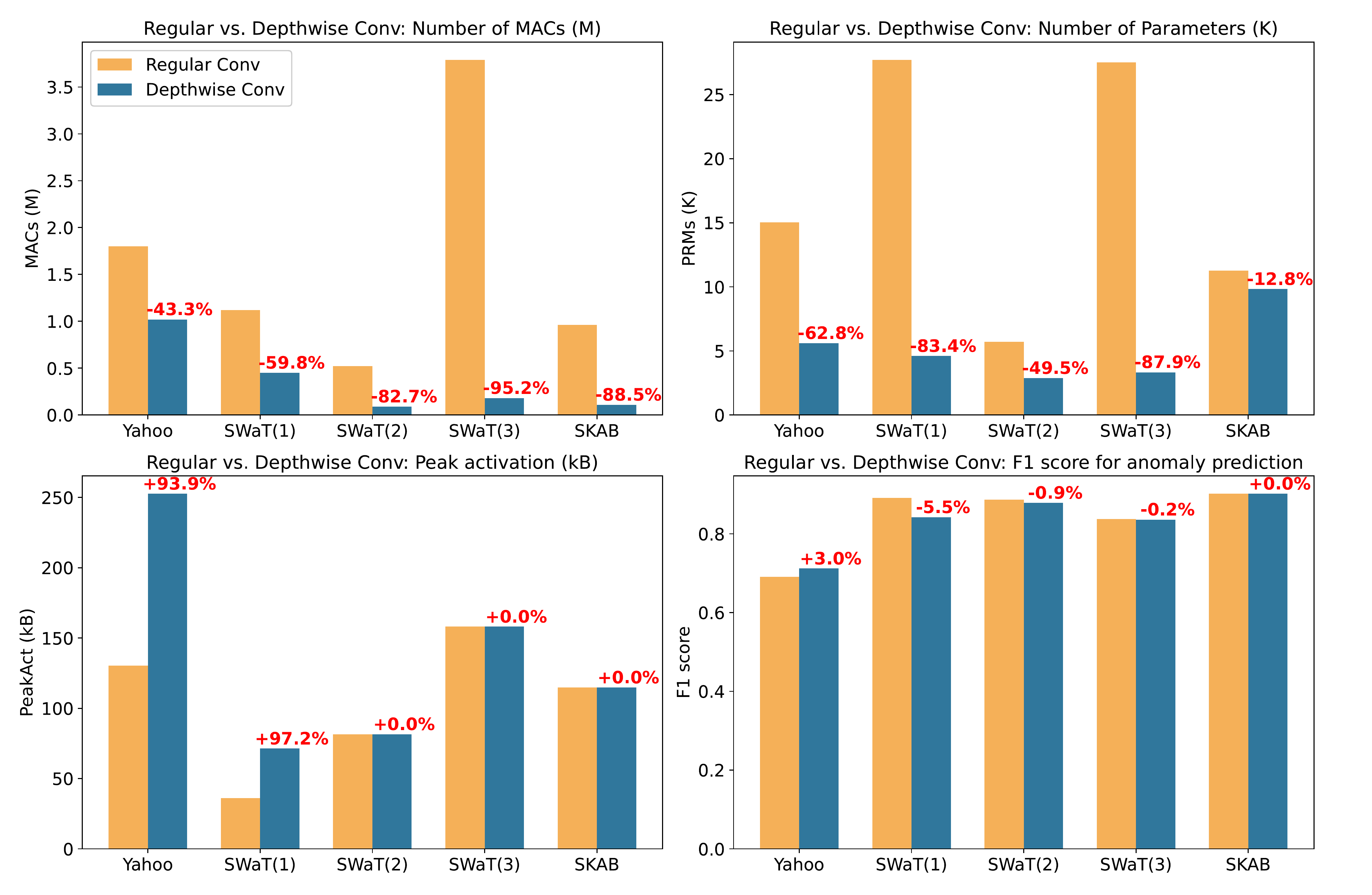}
  \caption{Analysis on five industrial datasets: comparison between regular convolution and depthwise separable convolution with respect to the number of MACs, number of parameters, peak activation size, and F1 score.}
  \label{fig:rg_dw_comparison}
\end{figure*}

In the temporal context, the most common approaches in the literature for anomaly detection are prediction-based models, which can determine whether a new data point is an anomaly upon arrival\cite{10.1145/3444690}. With the advent of deep learning, Convolutional Neural Network (CNN) and Long Short-Term Memory (LSTM) based prediction models emerge as the most powerful solution to anomaly detection with the SOTA accuracy\cite{8233155, 9748023}. Compared with LSTM, CNN shows its superiority in its model expressivity and training efficiency. Specifically, CNN-based temporal models can efficiently capture longer time scales by expanding larger receptive fields, but LSTM needs to propagate temporal information in a recurrent manner which leads to slower inference speed and larger memory cost. Moreover, CNN achieves remarkable performance in extracting time-independent and informative features from multi-dimensional data types by adjusting the dimension of convolution filters\cite{TII.2021.3069849, asoc.2018.04.024}. In contrast, LSTM is limited to input dimensions and cannot sufficiently capture time-independent features from high-dimensional inputs. Also, the layer-by-layer structure and kernel-wise operation of CNN models pave the way for memory rescheduling, while the stacked recurrent layers in LSTM impede the rescheduling of the deterministic hidden states. 
Motivated by the prominent characteristics of CNN, we first investigate the performance of CNN-based models for time series anomaly detection. The widely used CNN models for anomaly detection consist of convolution, pooling layers, and fully connected layers\cite{2018.2886457, 9748023}. However, the regular convolution operation faces expensive computation and storage costs given its nature of cross-channel computation. Unlike regular convolution, the depthwise separable convolution performs spatial convolution independently for each input channel and creates linear output combinations by pointwise convolution. We conduct a comprehensive empirical comparison of CNN models with regular convolution and depthwise separable convolution, including prediction accuracy (measured by F1 score), computation complexity (measured by the number of multiply-accumulate operations (MACs)), model size (measured by the number of parameters), and peak memory size (measured by peak activation size).  As shown in Fig.\ref{fig:rg_dw_comparison}, the depthwise separable convolution can significantly reduce the number of MACs and model parameters by 50-90\%, compared with the regular convolution. Meanwhile, the F1 score on anomaly detection tasks indicates that the depthwise separable convolution achieves comparable  prediction accuracy. \textcolor{black}{To retain the same performance as the regular convolution, the depthwise convolution is likely to double the number of channels. However, depthwise convolution still involves much fewer parameters compared to regular convolution, which weakens the model’s ability to learn strong representations from the data. Thus, we observe the slight performance drop (5.5\%) of dataset SWaT(1) utilizing depthwise convolution.} Although the depthwise separable convolution largely brings down the computation complexity and the model size with negligible accuracy loss, it still yields excessively high peak memory (i.e., peak activation size) for MCUs.

To reduce the peak memory of CNN models, MCUNet is proposed to reschedule the memory distribution among convolutional layers\cite{3496706, mcuv2}, and it is designed for the image classification setting that requires many convolutional layers in the CNN models. When it comes to the signal data or time series data, only a few (e.g., 1 or 2)  convolutional layers will suffice, and hence rescheduling the memory distribution does not work in this setting. Therefore, in this paper, we propose a novel framework Tiny Anomaly Detection (TinyAD), to significantly decrease the peak memory of CNN models for real-time anomaly detection on MCU devices. 

In this framework, we jointly optimize the peak memory consumption within and across the convolutional layers by innovatively  exploring and integrating two memory rescheduling strategies: in-place memory rescheduling within depthwise convolutional layers and patch-by-patch execution order across convolutional layers. Given the layer-by-layer execution of CNN-based models, the peak memory consumption is normally dominated by the layer with the largest size of activations. To reduce the activation size of the depthwise convolutional layer, a temporary buffer is allocated in SRAM to transfer and update the activation results between channels, rather than holding all input and output activations in memory. However, the in-place memory rescheduling strategy cannot reschedule the computation memory across convolutional layers. To further reduce the peak memory (i.e., peak activation size),  a patch-by-patch execution order across convolutional layers is proposed  to slice the feature map into small spatial regions and execute a small patch at a time instead of computing the whole activation. Nevertheless, this sequential patch execution comes at the price of a slower inference speed. To accelerate on-device inference (i.e., prediction), we employ image-to-column (im2col) \cite{im2col} to convert the kernel operation to matrix multiplication. We conduct extensive experiments on multiple real-world  datasets, and the experimental results show that our proposed TinyAD significantly decreases the peak memory consumption by 2-5 times, regardless of various peak memory-dominant layers. Although the patch-by-patch execution confronts extra computation caused by overlapped receptive field, the computation overhead is trivial as CNN models just need a limited number of layers for time series data. Also, as both strategies only manipulate the memory distribution during the inference process, our proposed TinyAD does not hurt the prediction accuracy of CNN models. 

The major contributions of this paper are as follows:
\begin{itemize}
\item To the best of our knowledge, this is the first work to unlock the possibility of performing anomaly detection tasks using SOTA CNN models on MCU devices without model compression. 
\item We perform a comprehensive analysis of depthwise separable CNNs and regular CNNs for anomaly detection and find that the depthwise separable convolution operation can significantly reduce the computation complexity (i.e., the number of MACs) and parameter size of CNN-based models by 50-90\%.
\item To largely reduce the peak memory cost of running CNN-based models, we propose a novel framework TinyAD, which integrates two complementary memory rescheduling strategies: in-place memory rescheduling within depthwise convolutional layers and patch-by-patch execution order across convolutional layers.
\item We conduct extensive experiments on multiple real-world datasets, and the experimental results show that our proposed TinyAD significantly decreases the peak memory of CNN-based models without compromising the prediction accuracy, paving the way for deploying CNN-based models on MCU devices.
\end{itemize}

\section{Preliminaries}
\label{sec:preliminaries}
In this section, we begin with the essential introduction of the depthwise separable convolution and signal decomposition methods that are used in this work.

\begin{figure}[hbt!]
  \centering
  \captionsetup{justification=centering}
  \includegraphics[width= \linewidth]{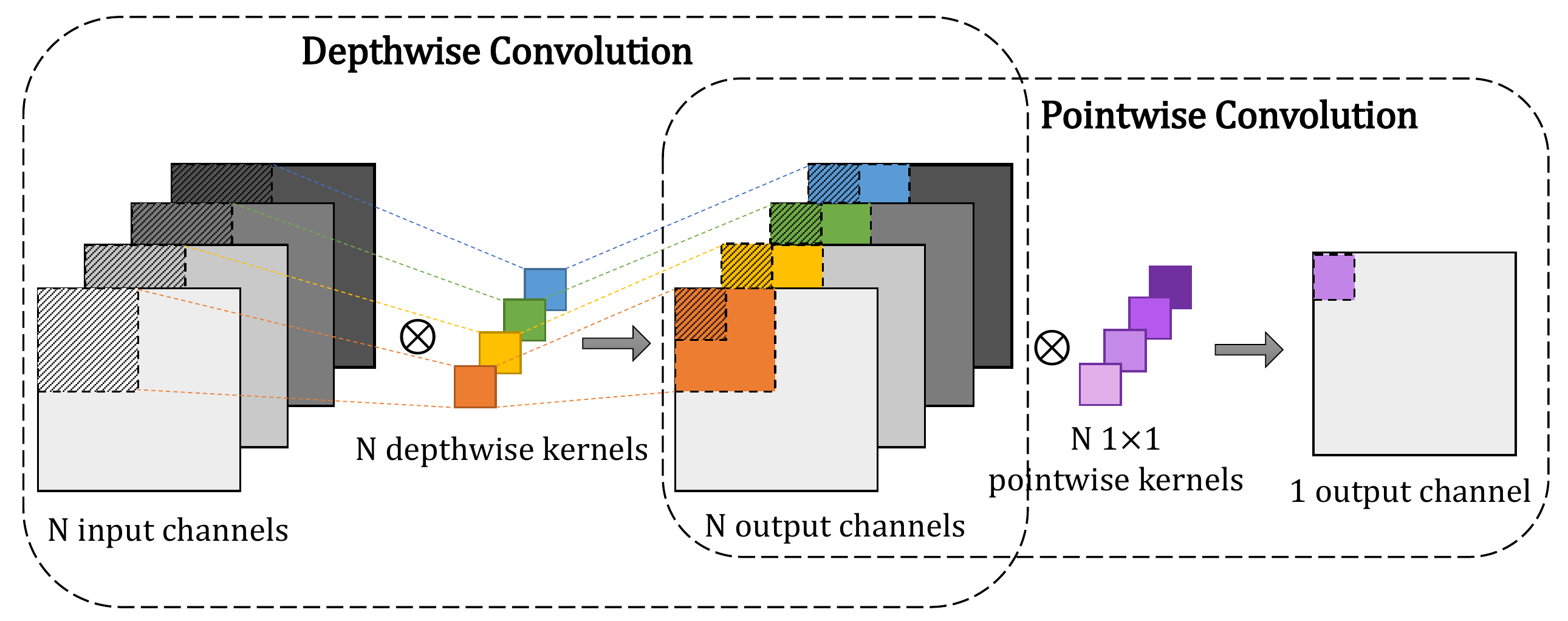}
  \caption{Architecture of depthwise separable convolution.}
  \label{fig:depthwise_cnn}
\end{figure}

\subsection{Depthwise separable convolution}
\label{subsec:dwcnn}
The depthwise separable convolution comprises a depthwise convolution and a pointwise convolution. The depthwise convolution performs channel-wise operations to model spatial relationships, while the pointwise convolution captures the cross-channel features by 1$\times$1 kernels. The factorized form of depthwise separable convolution is shown in Fig.\ref{fig:depthwise_cnn}. 

Formally, for a two-dimensional CNN (2D-CNN), we can represent a regular convolutional layer as a 4D tensor $\mathbf{C} \in \mathbb{R}^{n_\mathrm{i} \times n_\mathrm{o} \times k_{\mathrm{h}} \times k_{\mathrm{w}}}$, where $n_{\mathrm{i}}$ and $n_{\mathrm{o}}$ denote the number of input and output channels, and the spatial height and width of kernels are represented as $k_{\mathrm{h}}$ and $k_{\mathrm{w}}$, respectively \cite{bmvc18}. Unlike regular 2D convolution that captures intra- and inter-channel correlations in one go, the depthwise separable convolution saves both computational and space complexity by dividing the traditional convolution into two steps. Specifically, the spatial feature-learning kernel of depthwise convolution is denoted as $\mathbf{D} \in \mathbb{R}^{n_{\mathrm{i}} \times K \times k_{\mathrm{h}} \times k_{\mathrm{w}}}$, where the positive integer $K$ is the depthwise multiplier that determines the number of output channels after depthwise convolution as $K\times n_{\mathrm{i}}$. Fig.\ref{fig:depthwise_cnn} demonstrates the standard depthwise convolution where $K = 1$ and $\mathbf{D} \in \mathbb{R}^{n_{\mathrm{i}} \times 1 \times k_{\mathrm{h}} \times k_{\mathrm{w}}}$, which indicates that the number of kernels is the same as the number of input channels. Also, the pointwise convolution kernel for channel combination is represented as $\mathbf{P} \in \mathbb{R}^{Kn_{\mathrm{i}} \times n_{\mathrm{o}}  \times 1 \times 1}$. For an input patch $\mathbf{x}\in \mathbb{R}^{n_{\mathrm{i}}\times k_{\mathrm{h}}\times k_{\mathrm{w}}}$, given the two 4D tensors $\mathbf{D}$ and $\mathbf{P}$, the output vector $\mathbf{y}$ after the depthwise separable convolution is
\begin{equation}
\label{eq1}
\mathbf{y} = \left(\mathbf{P} \circ \mathbf{D} \right) \ast \mathbf{x}
\end{equation}
where $\circ$ is the compound operation, $\ast$ is the convolution operation. Given a kernel for depthwise convolution as $\mathbf{D}_{i,k} =\mathbf{D}[i,k,:,:]\in \mathbb{R}^{k_{\mathrm{h}} \times k_{\mathrm{w}}}$, and a kernel for pointwise convolution as $\mathbf{P}_{j,o}=\mathbf{P}[j,o,:,:]\in \mathbb{R}^{1\times 1}$. Each entry in $\mathbf{y}$ is obtained via $y_o=\sum_{k=1}^{K}\sum_{i=1}^{n_\mathrm{i}} \mathbf{P}_{{{n_\mathrm{i}}\times\left(k-1\right)+i},o} \ast \left(\mathbf{D}_{i,k} \ast \mathbf{x}_i \right)$.

Considering a feature map of size $H\times W$, the computational complexity of regular convolution is $O\left(H \times W \times n_{\mathrm{i}} \times n_{\mathrm{o}} \times k_{\mathrm{h}} \times k_{\mathrm{w}}\right)$, and the number of model parameters is given by $N = n_{\mathrm{i}} \times n_{\mathrm{o}} \times k_{\mathrm{h}} \times k_{\mathrm{w}}$. In contrast, depthwise separable convolution is more efficient in computation by achieving the computational complexity of $O\left(H \times W  \times n_{\mathrm{o}} \times K \times \left(n_{\mathrm{i}}  + k_{\mathrm{h}} \times k_{\mathrm{w}}\right)\right)$. Furthermore, the number of model parameters is significantly reduced to $N = K \times n_{\mathrm{i}} \times \left( n_{\mathrm{o}} + k_{\mathrm{h}} \times k_{\mathrm{w}}\right)$.

\subsection{Signal decomposition methods}
\label{subsec:signal}
To acquire supplementary information from time series data, feature extraction is one of the key procedures that can directly affect anomaly detection accuracy. Therefore, we introduce the methods used for extracting time, frequency, and time-frequency domain features from raw time series input \cite{TIE.2017.2733438}.

\subsubsection{Time features}
The raw time series data are intrinsically represented in the time domain. As temporal feature extraction methods aim to capture the morphology of time series, statistical methods are directly applied to capture the traits in the time domain. Considering a signal $\mathbf{z} = \{z_1, z_2, z_3, ..., z_n\}$, we explored 12 time domain features as described in Table \ref{tab:td}.

\begin{table}
\small
\centering
  \begin{tabular}{cc}
    \toprule
    \textbf{Quantity} & \textbf{Equations} \\
    \midrule
    Min & $ \mathrm{min} =  \mathrm{min}\left (\mathbf{z}\right)$\\[3pt]
    Mean & $\mu = \frac{1}{n}\sum_{i=1}^{n}z_i$\\[5pt]
    Root mean square & $ \mathrm{rms} = \sqrt{\frac{1}{n}\sum_{i}z_i^2}$\\[5pt]
    Variance & $ \mathrm{var} = \frac{1}{n-1}\sum_{i=1}^{n}{\left(z_i-\mu\right)}^2$\\[5pt]
    Standard deviation & $ \mathrm{std}\left(\sigma\right) = \sqrt{\frac{1}{n-1}\sum_{i=1}^{n}{\left(z_i-\mu\right)}^2}$\\[5pt]
    Peak & $ \mathrm{p} =  \mathrm{max}\left(\lvert \mathbf{z} \rvert\right)$ \\[3pt]
    Peak-to-peak & $ \mathrm{p2p} = \mathrm{max}\left (\mathbf{z}\right)- \mathrm{min}\left (\mathbf{z}\right)$\\ [3pt]
    Crest factor & $\mathrm{cf} = \frac{\mathrm{p}}{\mathrm{rms}}$\\[5pt]
    Skewness & $\mathrm{skew} = \frac{\frac{1}{n}\sum_{i=1}^{n}\left(z_i-\mu\right)^3}{\sigma^3}$\\[5pt]
    Kurtosis & $\mathrm{kurt} = \frac{\frac{1}{n}\sum_{i=1}^{n}\left(z_i-\mu\right)^4}{\sigma^4}$ \\[5pt]
    Form factor & $\mathrm{ff} = \frac{\sqrt{\frac{1}{n}\sum_{i=1}^{n}z_i^2}}{\mu}$\\[5pt]
    Pulse indicator & $\mathrm{pi} = \frac{\mathrm{p}}{\mu}$\\[3pt]
    \bottomrule
  \end{tabular}
\captionsetup{justification=centering}
\caption{Handcrafted time domain feature sets.}
\label{tab:td}
\end{table}

\subsubsection{Frequency features}
Fourier transform is a widely used method to convert the time-domain representation of time series to the power spectral density, which describes the relative magnitudes of a time signal against its frequency components. Given a frequency $f_i$, the signal power is represented as $S\left(f_i\right)$. We considered four frequency domain features which are described in Table \ref{tab:fd}.

\begin{table}
\small
\centering
  \begin{tabular}{cc}
    \toprule
    \textbf{Quantity} & \textbf{Equations} \\
    \midrule
    Spectral power & $\mathrm{sp} = \sum_{i=1}^{k}\left(f_{i}\right)^{3} S\left(f_{i}\right)$ \\[5pt]
    Mean power frequency & $\mathrm{mpf} = \frac{1}{k}\sum_{i=1}^{k}\frac{f_i S(f_i)}{\sum_{i=1}^{k}S(f_i)}$\\[7pt]
    Spectral skewness & $\mathrm{sskew} = \sum_{i=1}^{k}\left(\frac{f_{i}-\bar{f}}{\sigma}\right)^{3} S\left(f_{i}\right)$\\[5pt]
    Spectral kurtosis & $\mathrm{skurt} = \sum_{i=1}^{k}\left(\frac{f_{i}-\bar{f}}{\sigma}\right)^{4} S\left(f_{i}\right)$\\[5pt]
    \bottomrule
  \end{tabular}
\captionsetup{justification=centering}
\caption{Handcrafted frequency domain feature sets.}
\label{tab:fd}
\end{table}

\subsubsection{Time-frequency features}
To analyze how the frequency components of a time series change over time, we conduct discrete wavelet transform (DWT) to decompose the time series into a mutually orthogonal set of wavelets\cite{TIE.2017.2733438}. By scaling and translating the mother wavelet $\psi\left(t\right)$ (e.g. Daubechies), the DWT is expressed as
\begin{equation}
\label{eq2}
\psi_{j, k}(t)=2^{-j / 2} \psi\left(2^{-j} t-k\right)
\end{equation}
where $k \in [1,2^{-j}N]$ is a location index and $N$ denotes the number of observations. $j \in [0,J]$, and $J$ is the number of scales. To quantify the time-frequency features, we obtain the coefficients of DWT at different levels as $\mathbf{w}_{\phi}(i)$, and the wavelet energy is given by $\sum_{i=1}^{N} \mathbf{w}_{\phi}^{2}(i)/N$. In this work, we explore the wavelet energy ranging from level 1 to level 3 generated by mother wavelet Daubechies 1 (db1) and Daubechies 2 (db2).

\section{Methodology}
\label{sec:methodology}
In this paper, we propose TinyAD, a framework that enables depthwise separable CNN-based anomaly detection model on MCU embedded edge devices. CNN is a typical sequential network architecture as the activation results are propagated layer by layer. Also, each receptive field can be identified and disjoined from the whole feature maps as a patch thanks to the independence of kernel operations. These specialties of CNN allow the flexibility of memory rescheduling during the forward pass. When deploying machine learning techniques on MCU devices, all model parameters and input data are stored in the Flash memory. At the same time, SRAM takes over all computations (e.g., convolution), holds and relieves the parameters and activation results alternatively during the inference process. 

\begin{figure}[hbt!]
  \centering
  \captionsetup{justification=centering}
  \includegraphics[width= 0.7\linewidth]{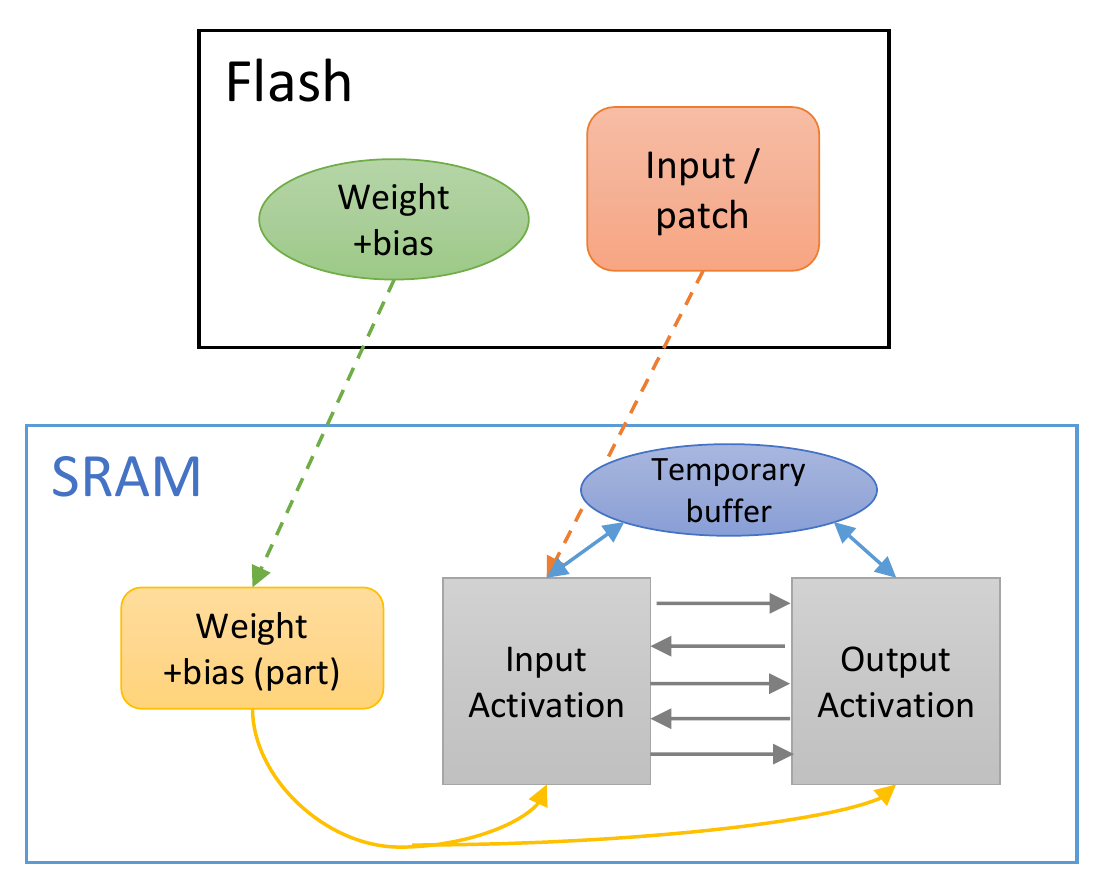}
  \caption{Memory allocation of CNN-based anomaly detection models on MCU.}
  \label{fig:mcu}
\end{figure}

The Flash and SRAM memory allocation for the CNN-based model is described in Fig.\ref{fig:mcu}. \textcolor{black}{In practice, the parameters (weights and biases) of each layer can be structured as a list of objects in a saved file (e.g., .json file), allowing us to load and parse one layer at a time, rather than load all model parameters in SRAM all at once. The memory space which holds parameters of the current layer will be released after the layer-specific computation finishes, and refilled by newly loaded parameters of the next layer. In TinyAD, the input patch is first loaded to the space in SRAM, and then the output from the lower layer will then serve as the input for the subsequent upper layer. 
Thus, we only need to allocate two variables/slots to store
the intermediate results exchangeably. The temporary buffer is designed specifically for the depthwise convolution layer to temporarily hold output activations derived from each input channel}, which is discussed in Section \ref{subsec:in-place}. In this memory allocation scheme, the computation results of each layer are not transferred back to the Flash until the inference process ends, which prevents the computation latency caused by the I/O stream. 

\begin{figure*}[hbt!]
  \centering
  \captionsetup{justification=centering}
  \includegraphics[width= \linewidth]{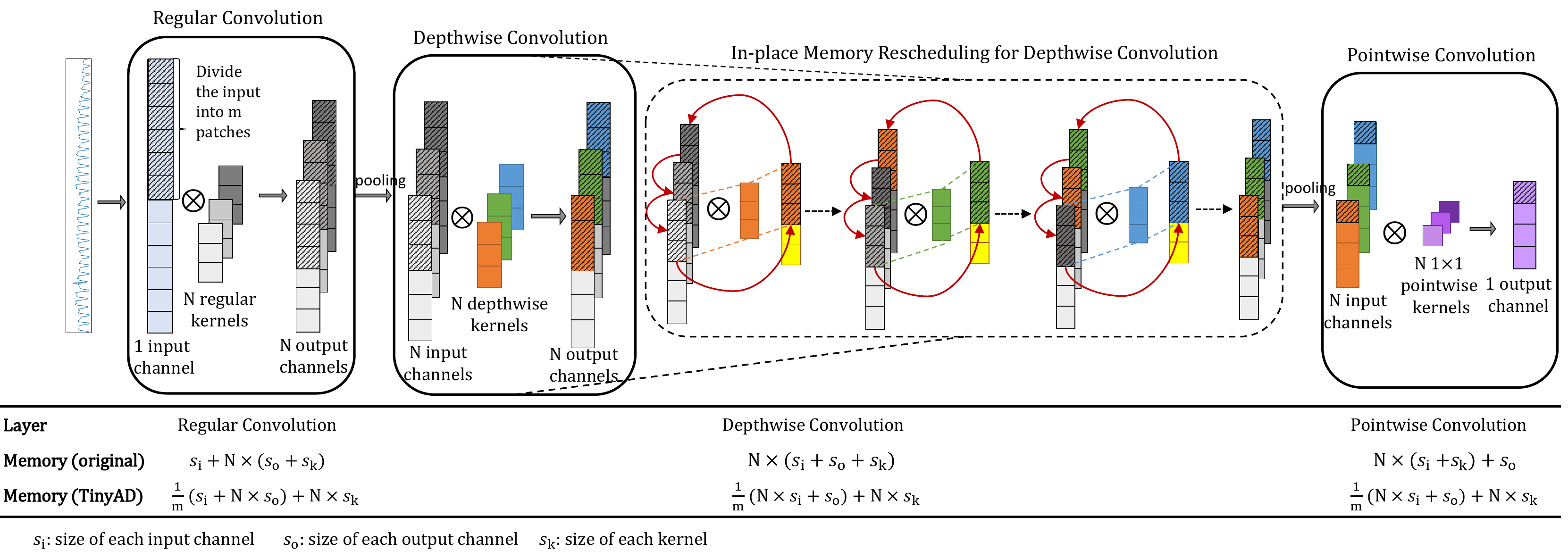}
  \caption{Architecture of TinyAD: patch-by-patch and in-place memory rescheduling methods for time series anomaly detection. (1D-CNN)}
  \label{fig:tinyAD}
\end{figure*}

\begin{figure*}[hbt!]
  \centering
  \captionsetup{justification=centering}
  \includegraphics[width= \linewidth]{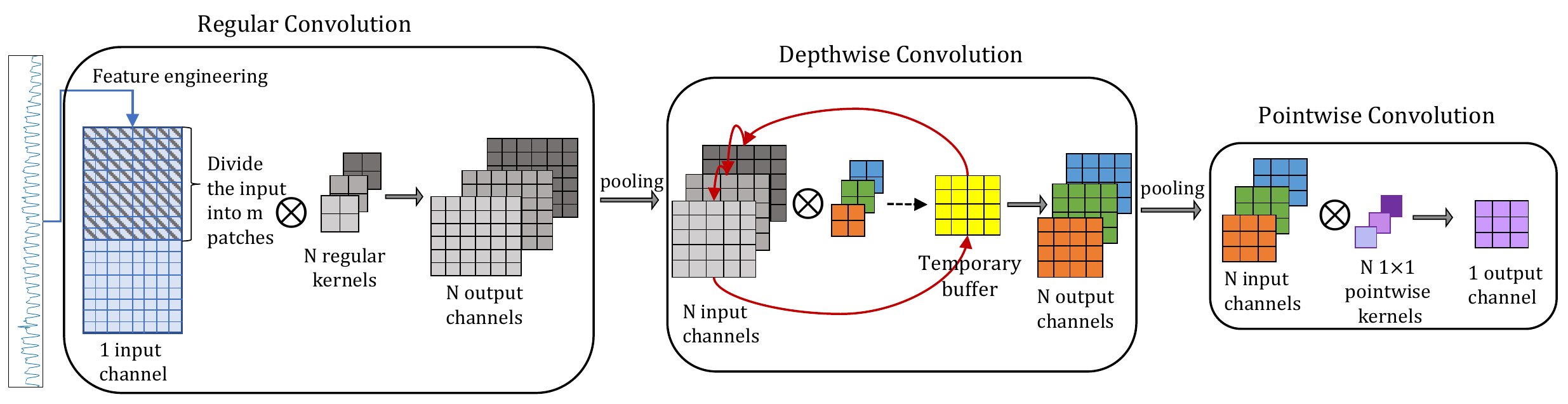}
  \caption{TinyAD for 2D-CNN.}
  \label{fig:tinyAD2d}
\end{figure*}

Fig.\ref{fig:tinyAD} illustrates a 1D-CNN example with raw time series as input. Without memory rescheduling techniques, the memory consumption of each convolutional layer is determined by the sum of input and output activation sizes across all channels. As a result, the peak memory is determined by the dominating layer with the largest activation size in total. However, the memory consumption of each layer can be largely reduced after we deploy the approaches of memory rescheduling, among which the memory consumption of a convolution layer is reduced by at most $2m$ times with the benefit of both patch-by-patch and in-place memory rescheduling methods, where $m$ is the number of patches we divide the input time series into. \textcolor{black}{For 2D time series input with multi-domain features, TinyAD is also applicable as we describe in Fig.\ref{fig:tinyAD2d}}

In this section, we first discuss the in-place memory rescheduling method specifically for depthwise convolution layers, and the patch-by-patch method to control the execution order of feature maps. Then, we demonstrate the benefits of combining these two methods, which further alleviates the memory limit of anomaly detection models regardless of the memory-dominating layer. 
\subsection{Depthwise convolution with in-place memory rescheduling}
\label{subsec:in-place}
\begin{figure}[hbt!]
  \centering
  \captionsetup{justification=centering}
  \includegraphics[width=\linewidth]{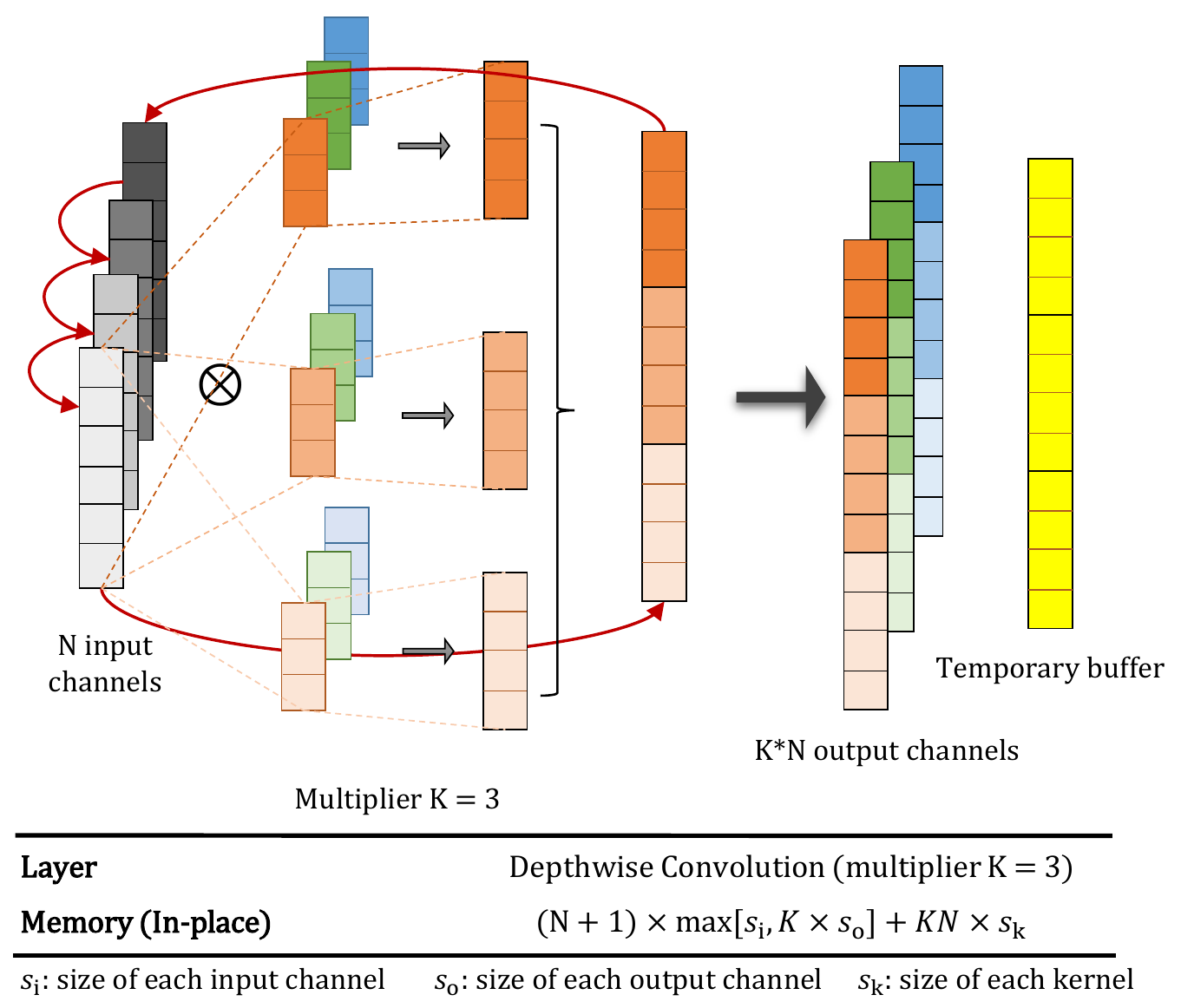}
  \caption{Example of in-place memory rescheduling when depthwise multiplier K = 3.}
  \label{fig:inplacek}
\end{figure}

As discussed in Section \ref{subsec:dwcnn}, depthwise separable convolution replaces the feature learning process performed by regular convolutions with two phases: an intra-channel spatial feature learning phase, and a cross-channel feature combination phase. Considering the convolution operations are conducted on each channel independently, we can reschedule the memory footprint by assigning a temporary buffer to hold and transfer the activation results between channels. As described in Fig.\ref{fig:tinyAD}, during the inference of depthwise convolution, rather than hold the whole input and output activation results, we overwrite the input activation of a channel with the output activation of another channel in the queue by utilizing the buffer as a `transit station', and shift the input activation across the channels without information loss. As shown in Fig.\ref{fig:inplacek}, given a depthwise convolution with multiplier $K$, the activation size of each input and output channel denoted as $s_\mathrm{i}$ and $s_\mathrm{o}$ respectively, and size of each kernel denoted as $s_\mathrm{k}$, \textcolor{black}{the size of temporary buffer is determined by $\max \left( s_\mathrm{i}, K \times s_\mathrm{o}\right)$. This design ensures sufficient memory allocation when overwriting the input activation of a channel with the output activation of another channel in the queue.} Thus, we can reduce the memory consumption of depthwise convolution from $N\times s_\mathrm{i} + NK \times s_\mathrm{o}+NK \times s_\mathrm{k}$ to $\left(N + 1\right) \times \max \left( s_\mathrm{i}, K \times s_\mathrm{o}\right) + KN \times s_\mathrm{k}$. When $K=1$, the method achieves its best effectiveness by reducing the memory consumption of depthwise convolution by around $2$ times.

\subsection{Patch-by-patch execution rescheduling}
\label{subsec:patch}
Regular CNN without memory rescheduling performs convolution layer-by-layer, which requires a full load of input and output activation during the inference process. However, the patch-by-patch method cuts down memory usage by dividing the input feature map into small spatial regions, which are executed under the pre-designed CNN architecture in order. Each output patch of the last conventional layer is held in SRAM until all the patches are passed through the convolutional layers. Given a pre-designed CNN-based anomaly detection model, the receptive field (striped areas in Fig.\ref{fig:tinyAD}) can be identified and matched layer-by-layer with respect to the final output beforehand. Although obtaining the non-overlapping output patches comes at the price of overlapping receptive fields and repeated computation overhead, the time series anomaly detection models superbly avoid perceptible computation latency by its distinction of predicting anomalies with a limited number of convolutional layers. The adoption of patch-by-patch can effectively diminish the memory usage of each layer by $m$ times, where $m$ is the number of patches that the input time series has been divided into.

\begin{figure*}[hbt!]
  \centering
  \captionsetup{justification=centering}
  \includegraphics[width=\linewidth]{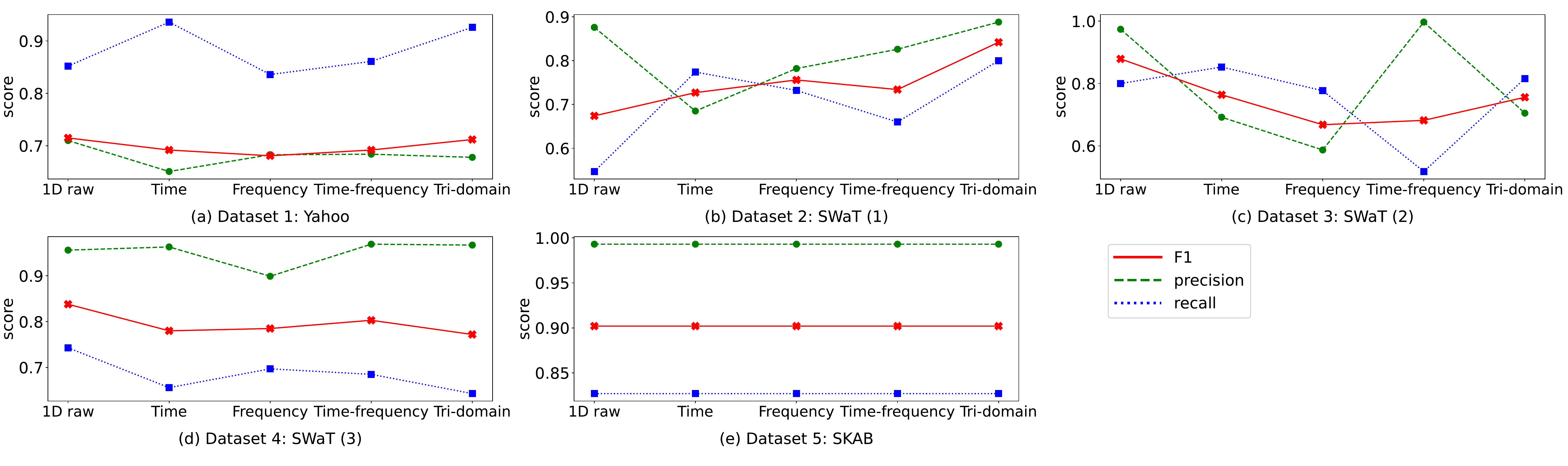}
  \caption{Experiment on depthwise separable CNN: impact of different time series features.}
  \label{fig:ts_features}
\end{figure*}
\subsection{Combining In-place and patch-by-patch schemes}
\label{subsec:comb}
As in-place and patch-by-patch methods reap the benefits of easing memory constraints respectively with intra-layer and inter-layer memory rescheduling, we combine these two strategies to maximize the potential of reducing the peak SRAM memory for CNN-based anomaly detection models. As described in Fig.\ref{fig:tinyAD}, each patch is passed through the layers and initiates the in-place memory rescheduling when executing the depthwise convolution. Although patch-by-patch is applicable to various types of convolutional layers, the in-place memory rescheduling can further reduce the memory consumption of the depthwise convolutional layer, which makes the whole process of convolution more memory efficient. When depthwise convolution dominates the peak memory consumption in SRAM, TinyAD can reduce the memory usage from $N\times s_\mathrm{i} + NK \times s_\mathrm{o}+NK \times s_\mathrm{k}$ to $\frac{\left(N + 1\right) \times \max \left( s_\mathrm{i}, K \times s_\mathrm{o}\right)}{m} + KN \times s_\mathrm{k}$. Specifically, for a standard depthwise convolution layer where the depthwise multiplier $K=1$, solely leveraging the patch-by-patch strategy can only reduce the memory by $m$ times, and in-place memory rescheduling can only reduce the memory by at most $2$ times. On the contrary, TinyAD can achieve its best performance by reducing the peak memory by $2m$ times. This improvement further addresses the memory bottleneck in anomaly detection models on MCU. 

\section{Experiment}
\label{sec:exp}
\subsection{Experiment settings} 
\label{subsec:expsetting}
\subsubsection{Dataset} 
We evaluate the CNN-based anomaly detection model on 4 industrial datasets and 1 commercial dataset to compare the memory efficiency of prediction tasks with and without the inference framework TinyAD. Each industrial dataset is a univariate sensor dataset, where the anomalies are abnormal sensor values caused by random or scheduled cyber attacks. The ratio of anomalies in SWaT is around 1\%, while the ratio of anomalies in SKAB is around 50\%. As an extension to verify that our framework not only works on industrial datasets but is also applicable to commercial datasets. We have included the dataset released by Yahoo Labs, where each time series contains around 0.2\%-1\% anomalies indicating outliers in computing systems. The decomposed signal features are extracted using the approaches introduced in Section \ref{subsec:signal}. A detailed description of each dataset is provided as follows: 
\begin{itemize}
\item \textbf{Secure Water Treatment testbed (SWaT)} \cite{71368-7_8}: SWaT is a scaled-down real-world industrial water treatment plant, which records network data across all sensors every second. In this work, we utilize three time series datasets from three sensors named `FIT 401', `LIT 101', and `LIT 301', which are marked as SWaT(1), SWaT(2), and SWaT(3) respectively in the experiments. These three datasets are all collected throughout 5-6 days with around 450,000 instances each.
\item \textbf{Skoltech Anomaly Benchmark (SKAB)} \cite{1693952}: SKAB is a miniature water circulation, control, and monitoring system, which consists of 34 multivariate time series collected by multiple sensors embedded on the testbed. After exploring the time series dataset, we extract the univariate sensor signal `RateRMS' over 5 hours, which has 18,161 instances to depict the circulation flow rate of the water inside the system. 
\item \textbf{Yahoo Webscope} \cite{yahoo}: This dataset is released by Yahoo Labs, which comprises 367 real and synthetic times series with labeled anomalies. In this work, we utilize the sub-dataset A1 benchmark as it consists of real-world time series, which depicts the aggregate status of Yahoo membership logins. Each time series contains around 1500 instances.
\end{itemize}

\begin{figure*}[t!]
  \centering
  \captionsetup{justification=centering}
  \includegraphics[width=0.8\linewidth]{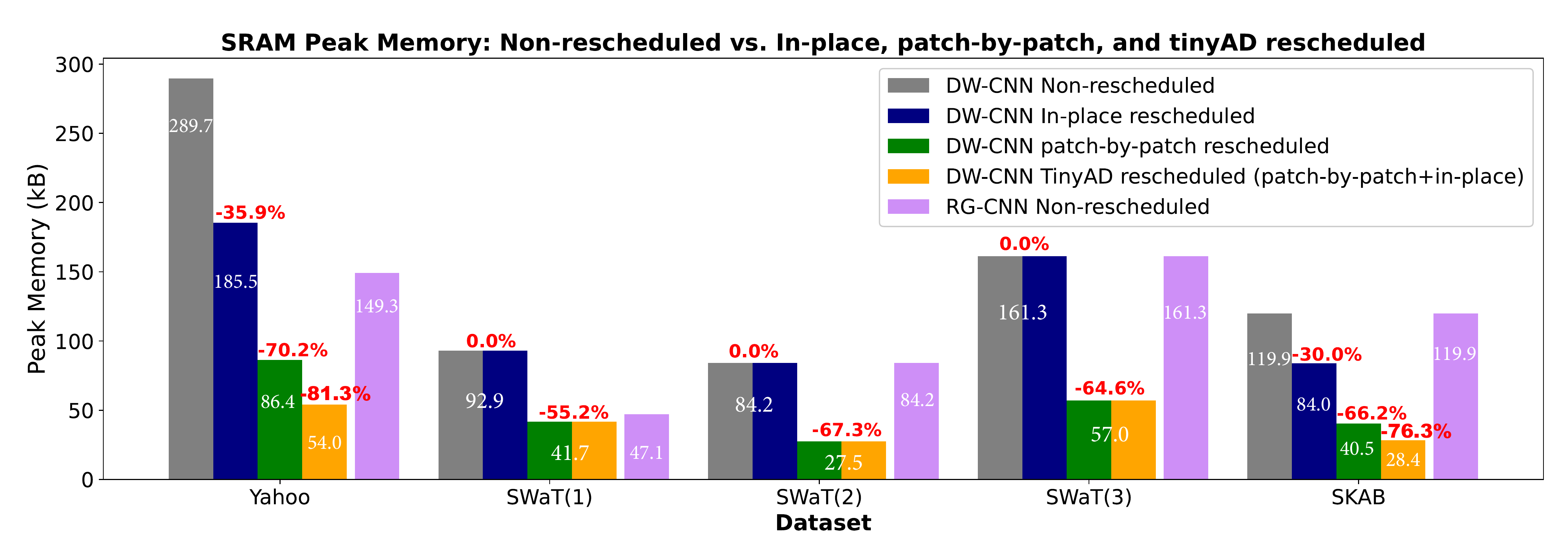}
  \caption{Comparison of memory efficiency under different memory rescheduling methods (number of patches = 3).}
  \label{fig:tinyad_pm}
\end{figure*}
\subsubsection{Parameter settings} 
In our work, we split each dataset into 60\% for training, 10\% for validation, and 30\% for testing. The hyper-parameters of models (e.g., kernel size, number of filters) are rigorously tuned to ensure the optimal performance of each model under each dataset as shown in Table \ref{tab:par}. As we discussed in Section \ref{sec:introduction}, the number of filters in depthwise convolution-based models is increased to remedy the accuracy loss, which leads to higher peak memory consumption compared to the regular CNN.


\begin{table}
\renewcommand{\arraystretch}{0.9}
\centering
  \begin{tabular}{cccc}
    \toprule
    \multirow{2}{*}{\centering \textbf{Dataset}} & \multirow{2}{*}{\centering \textbf{Kernel Size}}
     & \multicolumn{2}{c}{\centering \textbf{Number of filters}}\\[1pt]
     & & \textbf{Regular Conv} & \textbf{Depthwise Conv}\\
    \midrule
    Yahoo & $\left(4,4\right)$ & 32 & 32\\[1pt]
    SWaT(1) & $\left(5,5\right)$ & 64 & 64\\[1pt]
    SWaT(2) & 3 & 32 & 64\\[1pt]
    SWaT(3) & 3 & 64 & 128\\[1pt]
    SKAB & 3 & 16 & 32\\
    \bottomrule
  \end{tabular}
\captionsetup{justification=centering}
\caption{Hyper-parameter setting in the experiments.}
\label{tab:par}
\end{table}

Prediction-based anomaly detection model normally takes a historical window of time series to capture the temporal dynamics and tags the predicted result as a deviant or normal event. Since the size of the time window can significantly impact the memory consumption of models, we simply choose a consistent time window for all models in the experiment to verify the efficiency of the memory rescheduling framework under the same input settings. For datasets SWaT and SKAB, we take the sensor records of the past 20 minutes for anomaly detection. As the sensor data arrives in seconds, the input length for anomaly detection models is $L = 1200$. Given limited instances in the Yahoo dataset, we restrict the input length to $L = 200$.

\subsection{Performance and memory-efficiency comparison} 
\label{subsec:comparison}
To inspect the effectiveness of TinyAD for inference memory rescheduling, we first discuss the necessity of extracting time series features, and the impact of the multi-domain time series features. Then, we illustrate the results of memory optimization by applying in-place, patch-by-patch, and the fused technique TinyAD accordingly. We conduct the experiment based on the same model structure as we mentioned in Fig.\ref{fig:tinyAD}, followed by two fully connected layers. We use 1D convolutional kernels to extract temporal dependency from 1D raw time series, and 2D convolutional kernels to capture dependency of both multi-domain features and temporal patterns from 2D time series features.   

\subsubsection{Impact of different time series features}
In this experiment, we compare the performance of anomaly detection by utilizing raw time series and multi-domain time series features as input, including time domain, frequency domain, time-frequency domain, and tri-domain features which is a combination of the above three. \textcolor{black}{As the result in Fig.\ref{fig:ts_features} shows}, different datasets can experience the advantages of different types of time series inputs. Particularly, 1D raw time series input can fulfill the expectation of accurate anomaly detection on four datasets out of five by achieving higher F1 scores. Meanwhile, tri-domain features outperform other types of input features in dataset Yahoo and SWaT(1) by achieving both higher scores of F1 and recall. \textcolor{black}{However, dataset SKAB shows insensitivity towards feature engineering in anomaly detection tasks, given its striking difference between anomalies and normal records under scheduled cyber attacks. Thus, the temporal patterns can be easily captured by deep learning models even with raw input.} Thereby, the following exploration of memory efficiency is based on the time series features which facilitate the optimal performance in anomaly detection tasks. We exploit tri-domain features in the anomaly detection model for datasets Yahoo A1 and SWaT(1), and 1D raw time series as input for the other datasets. As a result, we verify the memory efficiency of different memory rescheduling methods for both 1D and 2D CNN models in the following sections.     

\subsubsection{memory efficiency of in-place memory rescheduling}
Fig.\ref{fig:tinyad_pm} shows the memory efficiency of depthwise separable convolution under different memory rescheduling methods. In-place scheduling can operate effectively for depthwise convolution-dominant models, as it aims to reduce the memory consumption during depthwise convolution, but is not applicable to regular convolutional layers. In our experiment, with depthwise multiplier $K=2$, in-place rescheduling can reduce the peak memory by 30\%-36\%, \textcolor{black}{but fails to reduce} the peak memory if it is dominated by regular convolutional layers.       

\subsubsection{memory efficiency of patch-by-patch memory rescheduling}
Patch-by-patch significantly breaks the memory bottleneck by reducing the peak memory to at least half as compared to the non-rescheduled model, regardless of the type of memory-dominant layers. In our experiment, we divide the input time series data into 3 patches and scale down the memory usage by 2-3 times solely with the patch-by-patch memory rescheduling method. 

\begin{table*}
\centering
\renewcommand{\arraystretch}{0.9}
  \begin{tabular}{ccccccc}
    \toprule
    \textbf{Dataset} & \textbf{Type} & \textbf{F1} & \textbf{MACs(M)} & \textbf{Model Size(kB)} & \textbf{PeakMem(kB)-float32} & \textbf{SRAM$\leq$64kB}\\
    \midrule
    \multirow{3}{0.15\textwidth}{\centering \textbf{Yahoo (2D CNN)}} & RG-CNN & 0.691	& 1.80 &	60.2 & 149.3 & \xmark \\
    & DW-CNN & 0.707 & 1.02 ($\downarrow$ 43\%) & 22.4 ($\downarrow \times$3) & 289.7 & \xmark \\
    & DW-CNN(TinyAD) & 0.707 & 1.02   & 22.4  &	54.0 ($\downarrow \times$5) & \cmark \\
    \midrule
    \multirow{3}{0.15\textwidth}{\centering \textbf{SWaT(1) (2D CNN)}} & RG-CNN & 0.891	& 1.12 & 110.8 & 47.1 & \cmark  \\
    & DW-CNN & 0.842 & 	0.45 ($\downarrow \times$2.5) &  18.4 ($\downarrow \times$5) & 92.9 & \xmark  \\
    & DW-CNN(TinyAD) & 0.842 & 0.45  & 	18.4  &	41.7 ($\downarrow \times$2) & \cmark  \\
    \midrule
    \multirow{3}{0.15\textwidth}{\centering \textbf{SWaT(2) (1D CNN)}} & RG-CNN & 0.887 & 	0.52 & 	22.8 & 	84.6 & \xmark  \\
    & DW-CNN & 0.879 &	0.09 ($\downarrow \times$6) &	11.5 ($\downarrow \times$2) &	84.6 & \xmark  \\
    & DW-CNN(TinyAD) & 0.879  & 0.09  & 11.5 & 27.5 ($\downarrow \times$3) & \cmark  \\
    \midrule
    \multirow{3}{0.15\textwidth}{\centering \textbf{SWaT(3) (1D CNN)}} & RG-CNN & 0.838 &	3.79 &	110.08 &	161.3 & \xmark  \\
    & DW-CNN & 0.836 & 	0.18 ($\downarrow \times$21) & 	13.3 ($\downarrow \times$9) &  161.3 & \xmark  \\
    & DW-CNN(TinyAD) & 0.836 & 0.18  & 	13.3  &	57.0 ($\downarrow \times$3) & \cmark  \\
    \midrule
    \multirow{3}{0.15\textwidth}{\centering \textbf{SKAB (1D CNN)}} & RG-CNN & 0.902 &	0.96 &	45.1 &	119.9 & \xmark  \\
    & DW-CNN & 0.902 &	0.11 ($\downarrow \times$9) &	39.3 ($\downarrow$ 12.8\%) &	119.9 & \xmark  \\
    & DW-CNN(TinyAD) & 0.902 & 0.11 & 39.3  & 28.4 ($\downarrow \times$4) & \cmark \\
    \bottomrule
  \end{tabular}
\captionsetup{justification=centering}
\caption{Comparison between regular CNN (RG-CNN), depthwise separable CNN (DW-CNN) with and without TinyAD for memory rescheduling (number of patches = 3).}
\label{tab:tinyad}
\end{table*}

\subsubsection{memory efficiency of TinyAD: combining patch-by-patch and in-place memory scheduling}
With the joint design of patch-by-patch and in-place method, the proposed TinyAD can further reduce the memory usage of depthwise convolution, while maintaining the reduction of regular convolution. Since TinyAD unlocks the full potential of memory rescheduling techniques, the peak memory usage is reduced by 2-5 times as compared to the non-scheduled model. \textcolor{black}{Based on the preliminary analysis of the memory distribution of layers, the memory bottleneck is normally given by the most memory-consuming convolutional layer, which has the largest sum of input and output activation sizes. On the one hand, if the memory bottleneck is determined by regular convolutional layers (e.g., on SWaT datasets), the overall memory reduction mainly relies on patch-by-patch memory rescheduling. Although the depthwise convolution layer still reaps benefits from the full model, the most memory-consuming layer (regular convolution) remains unaffected in this case. Thus, as we can observe in Fig.\ref{fig:tinyad_pm}, TinyAD conducted on dataset SWaT achieves the same memory efficiency as the patch-by-patch-only method.} On the other hand, for the depthwise convolution-dominant models (e.g., on Yahoo and SKAB dataset), the peak memory is further reduced by 37.7\% and 29.8\% respectively when compared with the patch-by-patch method, and 70.8\% and 66.1\% respectively when compared with the in-place method.

Table \ref{tab:tinyad} exhibits the overall performance and memory efficiency of regular CNN and depthwise separable CNN models for anomaly detection tasks. It is evident that using depthwise separable CNN models assisted with TinyAD can reduce the model size by 3-9 times, as well as peak memory usage by 2-5 times. After deploying TinyAD for depthwise separable models, all models become applicable on NodeMCU embedded devices (Flash memory $\leq$1MB and SRAM $\leq$64kB). Furthermore, since the MACs are significantly reduced, the inference process is accelerated as compared with regular CNN models. Besides, there is no visible increase of MACs for depthwise separable models when executing patch-by-patch, which indicates negligible computation latency caused by TinyAD. This computation efficiency profits from the pre-evaluation of receptive fields with minimum overlaps and \textcolor{black}{the limited convolutional layers} of anomaly detection models. 

\textcolor{black}{In our experiments, we evenly divided the input time series into 3 patches to explore the effectiveness of TinyAD. The optimal choice of patch size varies under different datasets and different model parameter settings. As we can observe from Table \ref{tab:tinyad}, dataset Yahoo ($\mathrm{patch\_size} = \left(60,20\right)$) and SWaT(3) ($\mathrm{patch\_size} = \left(421,1\right)$) achieved optimal patch size under the SRAM budget of 64kB. Although a smaller patch size can dramatically reduce peak memory usage, it also introduces more iterations of loading the parameters of each convolutional layer, as all patches’ outputs need to be stitched together to obtain the full outputs. Thus, to balance the inference latency and memory budgets, the best patch size is the maximum patch size that can fit in the given SRAM memory budget of MCU.}

\begin{table*}
\centering
\renewcommand{\arraystretch}{1.3}
\begin{tabular}{ccccc}
\toprule
\textbf{Dataset} & \textbf{Method} & \textbf{Total inference (ms)} & \textbf{Data preparation (ms)} & \textbf{Forward (ms)} \\
\hline
\multirow{3}{5em}{\centering \textbf{Yahoo}} & On-CPU (without TinyAD) & 22.06 & 2.58 &	15.84 \\
\cline{2-5}
& MCU simulation (single-thread) & 110.77 & \multirow{2}{*}{56.19} & \multirow{2}{*}{48.16} \\
\cline{2-3}
& MCU simulation (multi-thread) & 73.88 (\textcolor{black}{$\downarrow$33\%}) &  & \\
\hline
\multirow{3}{5em}{\centering \textbf{SWaT(1)}} & On-CPU (without TinyAD) & 9.54 & 2.21 &	3.74 \\
\cline{2-5}
& MCU simulation (single-thread) & 87.45 & \multirow{2}{*}{49.33} & \multirow{2}{*}{31.45} \\
\cline{2-3}
& MCU simulation (multi-thread) & 65.64 (\textcolor{black}{$\downarrow$25\%}) &  & \\
\hline
\multirow{3}{5em}{\centering \textbf{SWaT(2)}} & On-CPU (without TinyAD) & 21.12 & 1.41 &	16.10 \\
\cline{2-5}
& MCU simulation (single-thread) & 52.01 & \multirow{2}{*}{29.33} & \multirow{2}{*}{15.42} \\
\cline{2-3}
& MCU simulation (multi-thread) & 39.83 (\textcolor{black}{$\downarrow$25\%}) &  & \\
\hline
\multirow{3}{5em}{\centering \textbf{SWaT(3)}} & On-CPU (without TinyAD) & 21.59 & 1.52 &	16.07 \\
\cline{2-5}
& MCU simulation (single-thread) & 76.42 & \multirow{2}{*}{34.40} & \multirow{2}{*}{35.70} \\
\cline{2-3}
& MCU simulation (multi-thread) & 47.12 (\textcolor{black}{$\downarrow$38\%}) &  & \\
\hline
\multirow{3}{5em}{\centering \textbf{SKAB}} & On-CPU (without TinyAD) & 23.76 & 4.13 &	16.03 \\
\cline{2-5}
& MCU simulation (single-thread) & 116.68 & \multirow{2}{*}{97.64} & \multirow{2}{*}{12.14} \\
\cline{2-3}
& MCU simulation (multi-thread) & 97.64 (\textcolor{black}{$\downarrow$16\%}) &  & \\
\bottomrule
\end{tabular}
\captionsetup{justification=centering}
\caption{Comparison of inference time between on-CPU and on-MCU simulation (number of patches = 3). The percentage marked with ``$\downarrow$'' indicates the efficiency gain from the multi-thread setting compared with the single-thread.}
\label{tab:inference}
\end{table*}

\begin{figure*}[t!]
  \centering
  \captionsetup{justification=centering}
  \includegraphics[width=\linewidth]{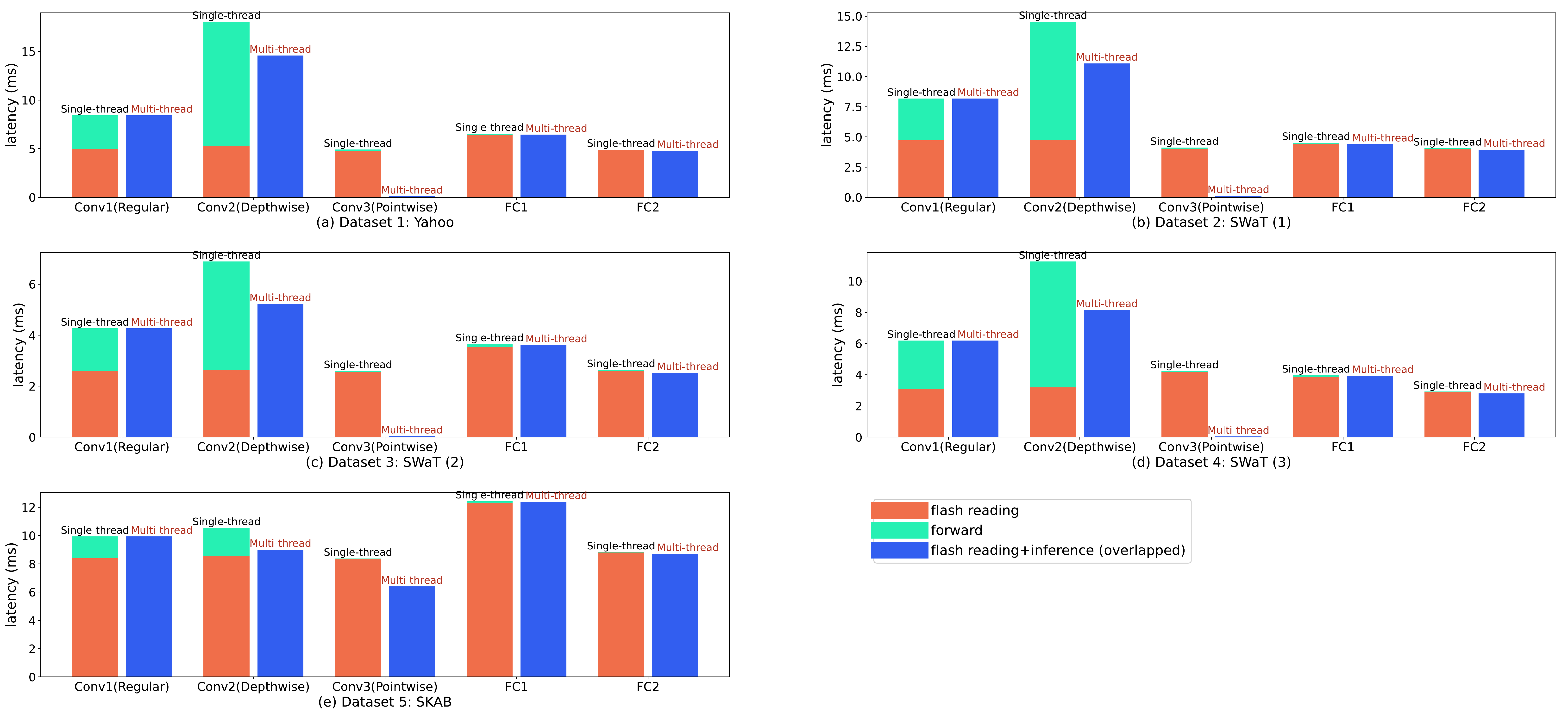}
  \caption{Inference latency of each layer (number of patches = 3). The y-axis measures the latency of both data loading and forward computation (each patch) for all convolutional layers, and the latency of all fully connected layers.}
  \label{fig:layer_lat}
\end{figure*}

\subsection{Comparison of inference latency} 
\label{subsec:comparison}
\textcolor{black}{Although TinyAD largely alleviates the memory bottleneck of deploying DNNs on MCUs, the inference time can be negatively impacted by the iterative flash-related data streaming and in-place rescheduling-related memory shift. Therefore, to estimate the time overhead during inference on MCU devices, we conduct experiments by simulating the workflow of DNN inference on MCU utilizing TinyAD, and compare the results with the on-CPU inference process without using TinyAD. For the MCU simulation of TinyAD framework, we explore the impact of data transmission between flash and SRAM under two settings: single-thread and multi-thread. The single-thread process has to load each layer and then perform layer-specific computation in a pipeline, while the multi-thread process allows loading the parameters of the next layer during the computation of the current layer.}

Table \ref{tab:inference} shows that the inference time is increased by the flash-related operational processes. It is worth mentioning that the on-CPU computation does not account for the potentially high communication cost from the data source (e.g., a sensor) to a computation node (e.g., a server) in an industrial scenario. As the result of patch-by-patch operation within TinyAD, the on-MCU simulation faces the iterative process of reading and decoding model parameters from the saved file onto the SRAM. Apart from this, rather than loading the whole file into SRAM at once, the flash-related operation latency mainly comes from retrieving the corresponding profile for each layer from flash before loading it to SRAM. Compared with the on-CPU full model, there is also a slight increase in the forward time caused by the in-place rescheduling approach. Notably, by utilizing multi-thread strategies for the on-MCU inference, the total inference time can be reduced by 16\%-38\%. 

\textcolor{black}{To better understand the benefits of introducing multi-thread processing during inference, we also explore the layer-wise inference latency, including the flash page read latency and other computations involved in the data loading process. The flash read latency $t_{\mathrm{read}}$ denotes the time the device needs to load a page of data for subsequent use after any read command is issued for a certain page address. Thus, the total read latency is  $N_{\mathrm{page}}\times t_{\mathrm{read}}$, where $N_{\mathrm{page}}$ is the number of retrieved pages.  In our experiments, $t_{\mathrm{read}} = 25\mu s$, and the page storage size is 8kB \cite{heidecker2010flash}. Fig.\ref{fig:layer_lat} shows the inference latency of each layer under both single-thread and multi-thread settings. As each layer only contains a small number of parameters that occupy only 1-2 units of flash page, the flash page read time is trivial compared to the overall data preparation process (3-12$ms$), which also includes the document decoding and parameter retrieval process. Generally, the multi-thread method can largely benefit both depthwise and pointwise convolutional layers since the flash-related operations can be processed simultaneously with the model forward computation. However, when the flash-related process dominates the execution duration, it becomes more challenging to reduce the inference time by utilizing multi-thread processing. Thus, experiments on the dataset SKAB show the smallest improvement, whereas SWaT(3) reaps the best results by completing most data reading tasks within the time of forward computation.} 

On average, the inference time of multi-thread TinyAD for on-MCU deployment is only 3 times slower than on-CPU deployment, showing a reasonable memory-speed trade-off where low memory consumption is prioritized for MCU. Besides, the inference time for all datasets using TinyAD is under 0.1s, which is considered tolerable for real-time anomaly detection tasks.

\section{Related work}
\label{sec:work}
\textbf{Anomaly detection for time series}. There are a variety of machine learning techniques that can identify anomalous points from the normal time series in an unsupervised and supervised manner. Supervised learning utilizes generative and discriminative models to find differences between abnormal and normal instances. However, the scarcity and unbalance of labels hinder the utilization of supervised methods \textcolor{black}{\cite{schmidl2022anomaly}}. Therefore, unsupervised learning is proposed to reconstruct the time-series data from historical records, and capture the deviation of reconstruction and ground truth as anomaly scores. CNNs and Recurrent neural networks (RNNs) are widely used as unsupervised anomaly detection techniques, in companion with autoencoder \cite{2018.2886457, Pankaj_Malhotra, TSMC.2020.2968516}. \textcolor{black}{Unlike RNN-based models, the attention mechanism shows its robustness in sequence modeling by parallel processing the series data. In sequential recommendation systems, the attention mechanism is widely used to extract relevant information from users' historical behaviors and predict their future preferences\cite{wang2019dmran, wang2019dmfp}.} Motivated by the effectiveness of self-attention in modeling long-term trends, an attention-based anomaly detection model is also proposed for anomaly detection, by utilizing multi-modal features extracted from broader temporal sequences\cite{3514067}. 

\textbf{Tiny machine learning}. The deployment of machine learning on IoT devices faces the challenges of limited computation and memory resources. Existing methods employ quantization\cite{CVPR.2019.00881} and channel pruning\cite{ICCV.2019.00339} to reduce the model size and floating-point operations (FLOPs). However, these strategies cannot alleviate the memory bottleneck induced by intermediate activation size, which forces the utilization of smaller models and low-fidelity inputs. To address this problem, MCUNet is proposed to reschedule the memory distribution of convolutional layers and shift the receptive fields to eliminate the computation overhead \cite{3496706, mcuv2}. \textcolor{black}{The experimental results of MCUNet reveal its significant improvement of memory efficiency on ResNet \cite{he2016deep} for image classification tasks. In time series, ResNet is also one of the promising frameworks for time series prediction and anomaly detection tasks. Given its CNN-based structure, our proposed TinyAD is also applicable to the model but requires additional
memory space to hold the input of residual blocks for identity mapping.}

Additionally, low power and energy consumption are other obstacles to deploying machine learning on edge devices. Thus, the power-efficient neural network implementation is proposed to relieve the power budget of the sensor node. \cite{3317829} jointly designed a hardware-oriented deep neural network (DNN) model with accelerators to reduce power consumption and accelerate inference computation. Also, Eciton\cite{FPL53798} is proposed for real-time inference of LSTM on low-power edge devices without accuracy loss. 

\textcolor{black}{\textbf{On-device model update}. Given the trend of deploying DNNs on edge devices, updating post-deployed DNN models on these memory-constrained devices is inevitable in future work. Given the costly communication and deficient label annotation in IoT systems, the most recent works explored the feasibility of updating the model on MCU devices by updating a compressed version of the model from server \cite{chen2022update}, and retraining the model on the device with streaming data \cite{lin2022device}. However, the problem of updating anomaly detection models on MCU devices is left unaddressed, which is well worth the investigation.}

\section{Conclusion}
\label{sec:conclusion}
In this paper, we propose TinyAD to reduce the peak memory consumption of anomaly detection models on IIoT devices. The experimental results prove the prediction effectiveness and memory efficiency of depthwise separable convolution, and illustrate that TinyAD can further optimize the memory distribution among convolutional layers based on the specific characteristics of convolutions. Experimental results show that our framework effectively reduces peak memory usage by 2-7 times with competitive prediction performance and smaller model size. As the implementation largely alleviates the memory budget of IIoT devices, it unseals the capability of real-time anomaly detection on low-cost sensor nodes.    

\section*{Acknowledgment}
We gratefully thank Dr. Thomas Taimre and Dr. Radislav Vaisman (the University of Queensland) for their valuable discussions and helpful comments on this research.

\bibliographystyle{IEEEtran}
\bibliography{citation}

\clearpage
\begin{IEEEbiography}[{\includegraphics[width=1in,height=1.25in,clip,keepaspectratio]{  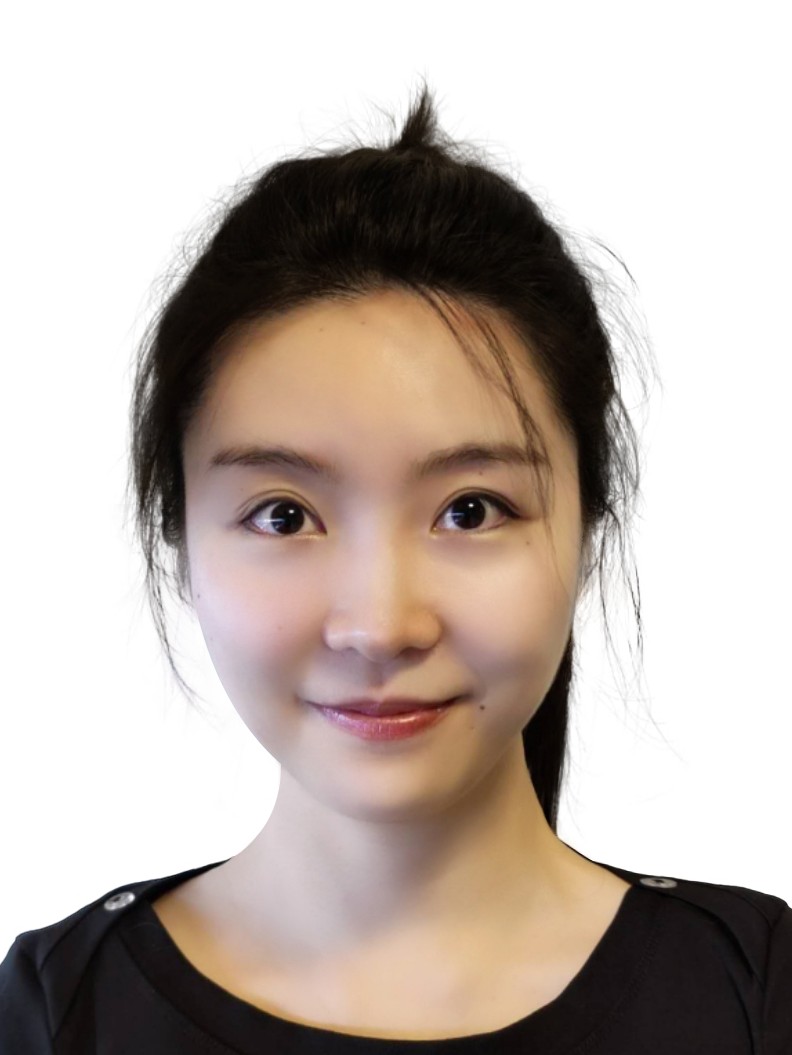}}]{Yuting Sun}
received her Master degree in
Data Science from The University of Queensland, Brisbane, QLD, Australia, in 2019. Currently, she is a Ph.D. candidate at the School of Information Technology and Electrical Engineering, the University of Queensland. Her research interests include data mining, time series anomaly detection, and spatiotemporal modeling.  
\end{IEEEbiography}

\begin{IEEEbiography}[{\includegraphics[width=1in,height=1.25in,clip,keepaspectratio]{  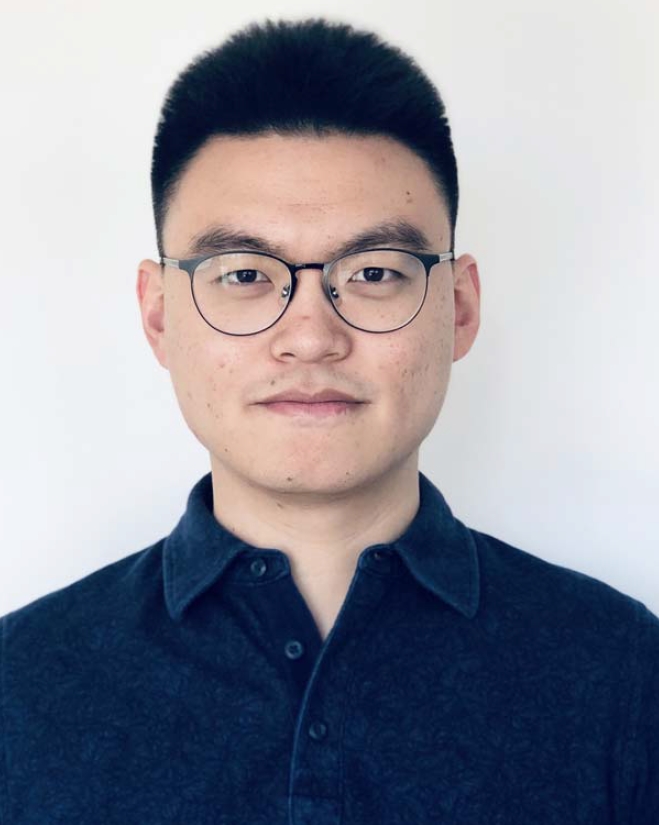}}]{Tong Chen}
received the Ph.D. degree in computer science from The University of Queensland, Brisbane, QLD, Australia, in 2020.
He is currently a Lecturer with the Data Science Research Group, School of Information Technology and Electrical Engineering, The University of Queensland. He is the recipient of the 2023 ARC Discovery Early Career Researcher Award (DECRA). His research interests include data mining, recommender systems, user behavior modeling, and predictive analytics. His research output includes over 60 publications, and has currently accumulated over 1,800 citations and an H-index of 25. Besides, he regularly serves as the program committee member of top conferences like KDD, SIGIR, and ICDM, as well as the invited reviewer of prestige journals like TKDE, TOIS, and TNNLS.
\end{IEEEbiography}

\begin{IEEEbiography}[{\includegraphics[width=1in,height=1.25in,clip,keepaspectratio]{  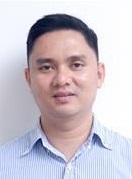}}]{Quoc Viet Hung Nguyen}
is a Senior Lecturer at Griffith University. He earned his Master and Ph.D. degrees from EPFL (Switzerland). He received the Australia Research Council Discovery Early-Career Researcher Award, in 2020. His research focuses on Data Integration, Data Quality, Information Retrieval, Trust Management, Recommender Systems, Machine Learning, and Big Data Visualization, with special emphasis on web data, social data, and sensor data.
\end{IEEEbiography}

\begin{IEEEbiography}[{\includegraphics[width=1in,height=1.25in,clip,keepaspectratio]{  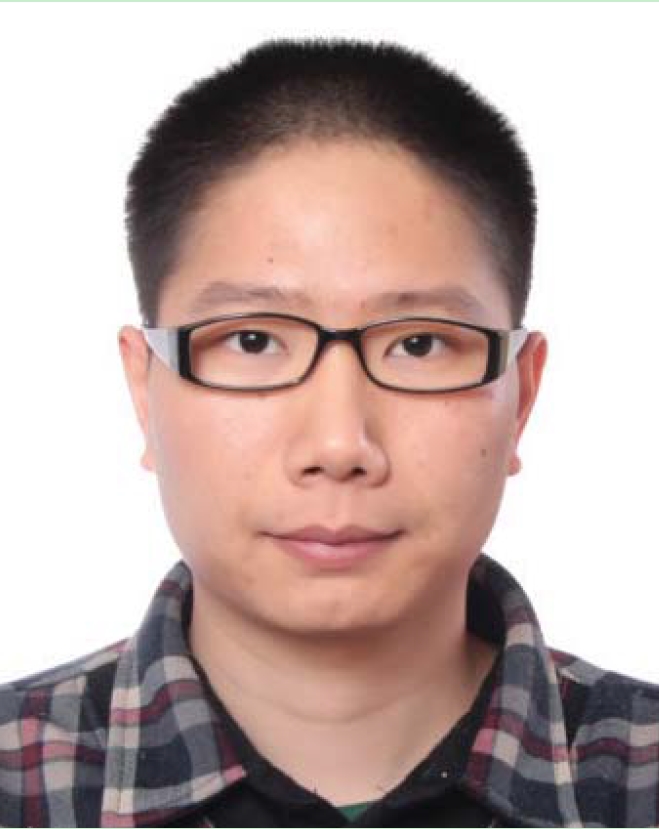}}]{Hongzhi Yin}
(Senior Member, IEEE) is an Associate Professor at The University of Queensland. He received his Ph.D.  from Peking University. He has made notable contributions to the fields of recommendation systems,  graph learning,  and edge intelligence and has received numerous awards and recognition for his research achievements. He has been named to IEEE Computer Society’s AI’s 10 to Watch 2022 and Field Leader of Data Mining and Analysis in the Australian Research 2020 magazine. In addition, he has received the Australian Research Council Future Fellowship 2022 and the Discovery Early Career Researcher Award (DECRA) 2016. He was featured among the 2022 and 2021 Stanford’s World’s Top 2\% Scientists.  He has published over 240 papers in the top conferences and prestigious journals and received H-Index 58. His research has won the Best Paper Award at ICDE 2019, Best Paper Award Runner-up at WSDM 2023, Best Student Paper Award at DASFAA 2020, Best Paper Award Nomination at ICDM 2018, and Peking University Distinguished Ph.D. Dissertation Award 2014.
\end{IEEEbiography}

\end{document}